%% file: main.tex
\documentclass[10pt,twocolumn,letterpaper]{article}

\usepackage{cvpr}              
\usepackage[accsupp]{axessibility}
\input{preamble}
\definecolor{cvprblue}{rgb}{0.21,0.49,0.74}
\usepackage[pagebackref,breaklinks,colorlinks,allcolors=cvprblue]{hyperref}
\usepackage{multirow}
\def\paperID{37908} 
\def\confName{CVPR}
\def\confYear{2026}

\title{ExpPortrait: Expressive Portrait Generation via Personalized Representation}

\author{
    Junyi Wang \quad Yudong Guo\thanks{Corresponding author.} \quad Boyang Guo \quad Shengming Yang \quad Juyong Zhang \\
    University of Science and Technology of China \\
    {\tt\small \url{https://ustc3dv.github.io/ExpPortrait/}}
}

\begin{document}

\twocolumn[{%
\maketitle
\begin{center}
  \includegraphics[page=1, width=1.0\textwidth, height=0.35\textheight]{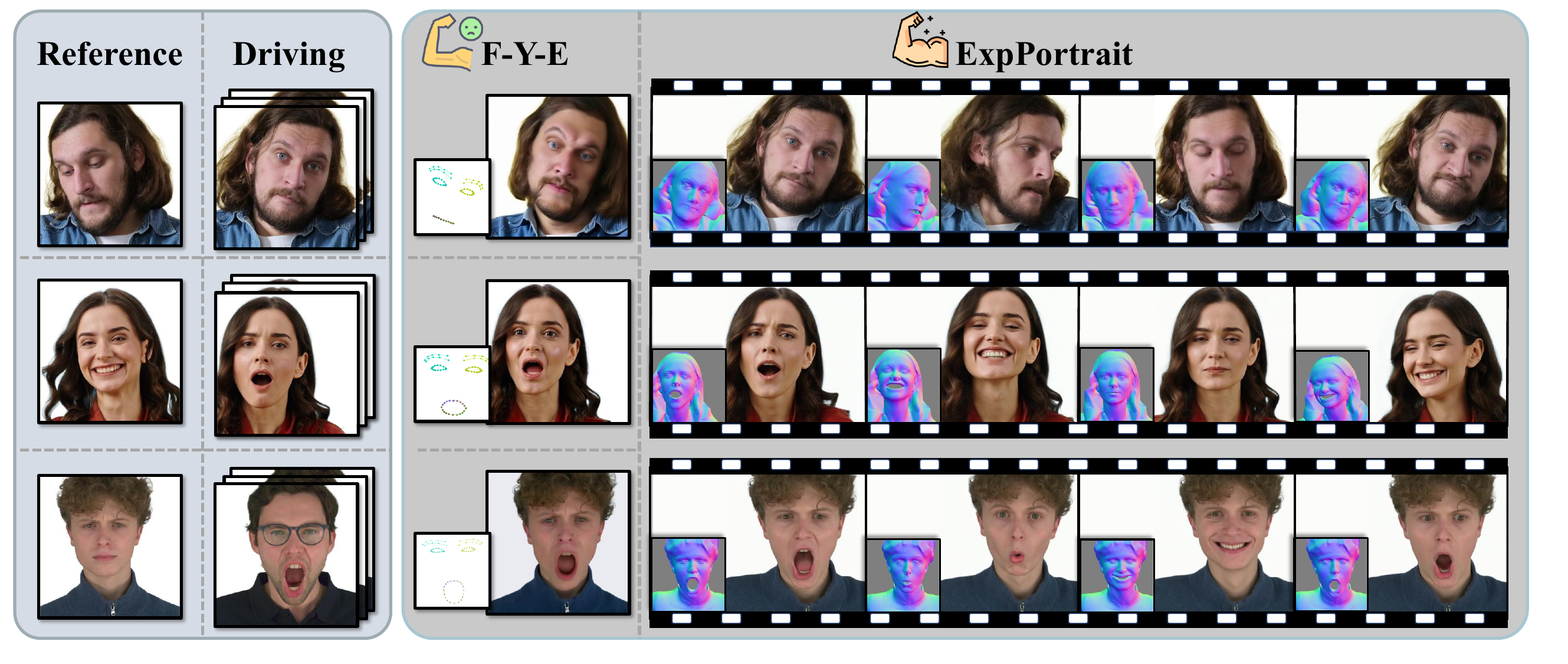}
   \captionof{figure}{ExpPortrait utilizes a personalized head representation for portrait animation, achieving video generation with high consistency and high-fidelity. This stands in contrast to methods like Follow-Your-Emoji~\cite{ma2024follow}, which are constrained by low-rank and smooth intermediate representations.}
  \label{fig:teaser}
\end{center}
}]
\insert\footins{\noindent\footnotesize $^*$Corresponding author.}

\input{sec/0_abstract}  
\input{sec/1_intro}
\input{sec/2_relatedwork}
\input{sec/3_method}
\input{sec/4_experiments}
\input{sec/5_conclusion}

\section*{Acknowledgments} This research was supported by the National Natural Science Foundation of China (No.62272433, No.62402468, No.U25A20390), Anhui Provincial Natural Science Foundation (No.2508085ZD011) and the Fundamental Research Funds for the Central Universities.
{
    \small
    \bibliographystyle{ieeenat_fullname}
    \bibliography{main}
}


\end{document}

%% file: sec/0_abstract.tex
\begin{abstract}
While diffusion models have shown great potential in portrait generation, generating expressive, coherent, and controllable cinematic portrait videos remains a significant challenge. Existing intermediate signals for portrait generation, such as 2D landmarks and parametric models, have limited disentanglement capabilities and cannot express personalized details due to their sparse or low-rank representation. Therefore, existing methods based on these models struggle to accurately preserve subject identity and expressions, hindering the generation of highly expressive portrait videos. To overcome these limitations, we propose a high-fidelity personalized head representation that more effectively disentangles expression and identity. This representation captures both static, subject-specific global geometry and dynamic, expression-related details. Furthermore, we introduce an expression transfer module to achieve personalized transfer of head pose and expression details between different identities. 
We use this sophisticated and highly expressive head model as a conditional signal to train a diffusion transformer (DiT)-based generator to synthesize richly detailed portrait videos. Extensive experiments on self- and cross-reenactment tasks demonstrate that our method outperforms previous models in terms of identity preservation, expression accuracy, and temporal stability, particularly in capturing fine-grained details of complex motion.
\end{abstract}

%% file: sec/1_intro.tex
\section{Introduction}
\label{intro}
Digital portrait generation is a key technology in computer vision and computer graphics, powering a wide range of applications from cinematic visual effects to lifelike virtual avatars in the metaverse. However, a major limitation of existing methods is the difficulty in achieving fine-grained, subtle expression control without compromising the consistency of the subject's identity. This paper aims to overcome this challenge by achieving precise control over subtle facial expressions while faithfully preserving identity.

Early controllable synthesis methods employed 2D landmarks~\cite{siarohin2019animating,Sagonas2013_300W,yang2023effective,TUFA,khirodkar2024sapiens} or parametric 3D models~\cite{pavlakos2019expressive,li2017learning,feng2021learning,li2023goha} as intermediate motion signals, but both suffer from severe representational limitations. 2D landmarks-based methods are limited by severe signal sparsity, lacking the geometric details needed to define subject identity or subtle micro-expressions, and exhibit poor stability under significant pose variations. 3D parametric models~\cite{li2017learning,pavlakos2019expressive} serve as low-rank, linear approximations of the human face. Their predefined blendshapes are insufficient to handle high-frequency nonlinear dynamics (e.g., wrinkles), leading to severe confusion between identity and expression. This fundamentally hinders the generation of high-fidelity, expressive portraits that preserve identity.

While the rapid development of diffusion models has markedly advanced portrait video generation~\cite{rombach2022high,guo2023animatediff,zhang2023dream}, existing methods still inherit the limitations of these flawed representations. Approaches using explicit 2D/3D parameters~\cite{chen2025echomimic,hu2024animate,peng2024controlnext,zhu2024champ} as control signals consequently struggle to capture high-frequency details or achieve accurate, identity-adaptive expression transfer. To address this, some studies~\cite{xu2025hunyuanportrait,xie2024x,zhao2025xnemo,cheng2025wan} have attempted to extract implicit motion features from the driving image via motion extractors, aiming to implicitly utilize learned motion information. Yet a major drawback is that the learned features are weakly controllable and insufficiently disentangled, leading to identity leakage and expression drift. Therefore, achieving robust identity preservation without compromising expressive fidelity remains an open and pressing problem.

As shown in \cref{fig:teaser}, given that existing parametric head models fail to provide a superior intermediate proxy for expressive portrait generation, we aim to improve the information density and controllability of this intermediate signal to better serve this task. Our goal is to construct a personalized head representation that captures unique identity structures and dynamic expression details with high fidelity. We first use SMPL-X~\cite{pavlakos2019expressive} mesh parameters as prior information. Based on this, we establish a disentangled relationship between identity and expression by optimizing two complementary offset fields: we learn a static per-vertex offset field for each subject to capture their unique high-frequency identity geometry. Meanwhile, we learn a dynamic per-vertex offset field designed to capture nonlinear skin deformations (e.g., wrinkles) corresponding to specific expressions. This constitutes a highly structured and detailed personalized head representation of identity and expression, significantly enhancing the expressive power of the head model.

However, applying this personalized, highly detailed head model to cross-identity expression transfer still faces compatibility challenges. The difficulty lies in how to adapt one subject's expression offset field to another subject's identity. We further construct an identity-adaptive expression transfer module to address this issue. This module takes the target subject's neutral identity mesh and target expression parameters as input. It employs a lightweight geometry MLP that predicts per-vertex dynamic offsets by conditioning on both the target's neutral geometry and an encoded representation of the driving signals. Finally, it outputs a mesh that presents the corresponding dynamic expression details, thus achieving vivid expression transfer without compromising identity. In summary, our contributions include the following aspects:
\begin{itemize}[leftmargin=*,noitemsep,nolistsep] \item We propose a personalized head representation with high disentanglement between identity and expression, capable of representing high-frequency geometric details and nonlinear facial deformations.

\item We design an identity-adaptive expression transfer module to address the inherent incompatibility problem of per-subject learning, enabling the accurate transfer of dynamic facial expression details to the target subject while preserving identity.

\item Based on the aforementioned personalized head model and expression transfer module, a diffusion generation model is trained, demonstrating state-of-the-art performance in identity preservation, expression richness, and controllability on both self- and cross-reenactment tasks. \end{itemize}

%% file: sec/2_relatedwork.tex
\begin{figure*}[ht]
\centering
\includegraphics[width=\linewidth]{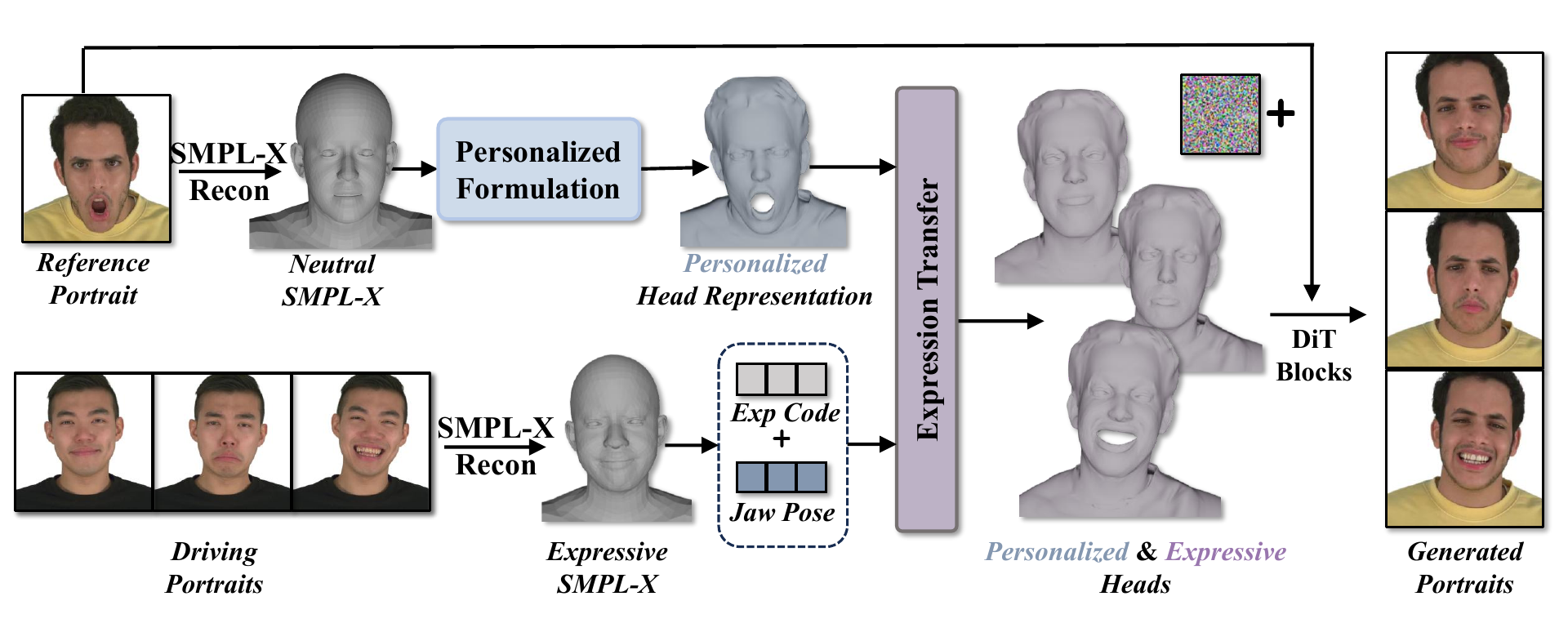}
\caption{Our framework. To address the limited decoupling capability and insufficient expressiveness of current parametric head representations, we propose a personalized head representation. Starting from the SMPL-X base mesh, we perform joint optimization learning of two complementary static and dynamic offset fields. We then construct an identity-adaptive expression transfer module to achieve cross-identity expression transfer. Using our head representation as a control signal, we guide a diffusion model for highly consistent and expressive portrait video generation.}
\label{fig:pipeline}
\end{figure*}

\section{Related Work}
\label{sec:related_work}

\noindent\textbf{Face Reenactment.}
Most face reenactment techniques~\cite{doukas2021headgan,pan2023drag,zhang2023metaportrait,guo2024liveportrait} utilize a Generative Adversarial Network (GAN)-based generator~\cite{goodfellow2020generative}, which uses facial cues or attributes, such as facial landmarks, as control signals. These approaches employ warping-based methods that attempt to learn implicit intermediate keypoints, which capture the facial motion between the source and target faces, thereby providing an alternative pathway for motion reenactment and transfer. However, these landmark representations tend to retain the target identity's facial structure, leading to identity drift and unnatural deformations when handling large pose differences or cross-identity reenactment. To improve geometric localization and reduce identity leakage, some studies have utilized 3D morphable parametric face models~\cite{blanz2003face,li2017learning,zielonka2022towards}, which can disentangle identity and expression, to preserve identity consistency across different subjects. Methods such as~\cite{feng2021learning,danvevcek2022emoca,qian2024gaussianavatars} enhance face models with detail displacements or emotion consistency losses to capture finer, detailed expressions. Other methods attempt to encode facial expressions into a latent vector and inject it into the generator network, or utilize Neural Radiance Fields (NeRF) and 3D Gaussians for head reconstruction and animation~\cite{guo2021ad,xiang2024flashavatar,mildenhall2020nerf,kerbl3Dgaussians,deng2024portrait4d}, but the consistency and rendering quality are still not fully satisfactory. The low-dimensional template subspace of the parametric face representations relied upon by these methods restricts their expressive power, hindering the capture of subject-specific anatomical structures or dynamic wrinkles and thus failing to achieve a sufficient balance between consistency and expressiveness.

\medskip
\noindent\textbf{Diffusion-based Portrait Generation.}
Diffusion models~\cite{ho2020denoising,guo2023animatediff,chen2026human} have demonstrated remarkable capabilities in generating diverse and high-fidelity results, significantly impacting portrait video generation. Within this domain, a primary challenge lies in effectively controlling the subject's motion while preserving their identity. A significant body of work conditions these diffusion models on explicit human representations to guide the generation. These control signals range from 2D sparse landmarks to rendered parameters from 3D parametric models. For instance, ~\cite{ma2024follow,chen2025echomimic,tu2025stableanimator,li2025dispose} utilize landmarks as spatial cues, while~\cite{wei2024aniportrait,wei2025magicportrait,qiu2025skyreels} employ 3DMM-derived maps to steer a latent diffusion model, prioritizing temporally consistent reenactment. While this explicit paradigm affords strong controllability, it is often constrained by the expressivity ceiling of the chosen representation. Sparse 2D controls can under-determine the full facial geometry and subtle expressions. Similarly, the low-rank subspaces of 3D templates struggle to model highly person-specific shapes or exaggerated expressions, leading to unnatural deformations. A parallel research thrust explores implicit motion control~\cite{zhao2025xnemo,gao2025constructing,xu2025hunyuanportrait}. These methods typically employ a motion encoder to extract driving dynamics directly from video frames, injecting these latent features into the diffusion backbone. While this implicit method can generate highly realistic results, it raises significant concerns regarding incomplete disentanglement. The learned latent motion features may inadvertently capture appearance cues from the driver, leading to identity leakage and compromising the separation of motion from appearance.

%% file: sec/3_method.tex
\section{Methodology}
\label{sec:method}

Given a reference portrait image $I_R$ and a driving video sequence $I_D = \{I_{D_i}\}_{i=1}^{F}$, our task is to generate a video $I_G = \{I_{G_i}\}_{i=1}^{F}$ where the subject from $I_R$ reenacts the poses and expressions of the subject in $I_D$, and $F$ is the number of frames. The result should preserve the reference identity with high fidelity while faithfully and expressively transferring the motion.

To establish a faithful mapping from the driving frames to the reference identity, we first build a personalized head representation that captures the subject’s identity and expression space (\cref{sec:personalized}). We then introduce an identity-dependent expression transfer module to robustly transfer poses and expressions across identities (\cref{sec:expression}). Finally, we fine-tune a pretrained video diffusion model~\cite{wan2025wan}, conditioning it on our personalized, detail-rich head representation to synthesize the final high-fidelity video (\cref{sec:diffusion}). The overall pipeline is shown in \cref{fig:pipeline}.

\subsection{Personalized Head Representation}
\label{sec:personalized}

Parametric representations, such as SMPL-X and FLAME, provide effective initializations for portrait generation and reenactment. Their blendshape structure promotes identity consistency and expressive control. However, as low-dimensional statistical models, they fail to capture high-frequency, subject-specific details. We therefore exploit SMPL-X’s 3D consistency by using its pre-tracked parameters as a coarse base layer, and augmenting them with decoupled, high-fidelity identity and expression details.


\subsubsection{Representation Formulation}
\begin{figure}[ht]
\centering
\includegraphics[width=\linewidth]{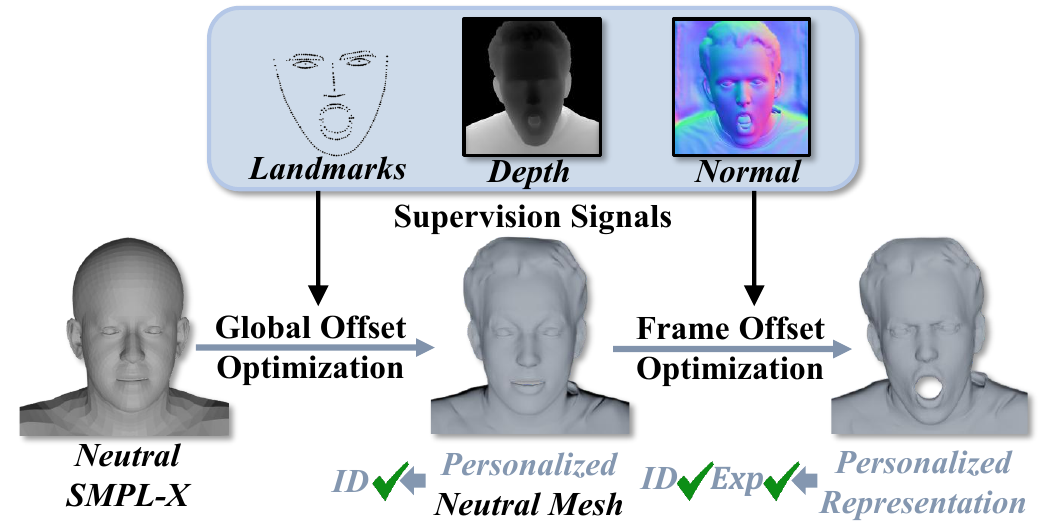}
\caption{An illustration of our optimization pipeline for transforming a generic SMPL-X mesh into our highly detailed, personalized head representation.}
\label{fig:Personalized_mesh}
\end{figure}

As shown in \cref{fig:Personalized_mesh}, our representation formulation process is as follows: Let $\mathcal{M}$ denote the function that returns mesh vertices given SMPL-X parameters. We first obtain a canonical neutral mesh for a given identity using only its shape parameters $\beta \in \mathbb{R}^{300}$, with the expression coefficients $\boldsymbol{\psi}$ and jaw poses $\boldsymbol{\omega}$ set to zero:
\begin{equation}
\label{eq:canonical_coarse}
V \;=\; \mathcal{M}(\beta)\big|_{\mathcal{P}} \;\in\; \mathbb{R}^{N\times 3},
\end{equation}
where $\mathcal{P}$ selects the head-and-shoulders vertex subset with $N$ vertices. Because this mesh is too sparse to capture person-specific geometry and subtle expressions, we upsample it to a dense mesh. Let $\mathcal{B}$ be a fixed barycentric interpolation operator that maps the coarse topology to a high-resolution topology with $N_s$ vertices ($N_s \gg N$):
\begin{equation}
\label{eq:canonical_fine}
V^{s} \;=\; \mathcal{B}(V) \;\in\; \mathbb{R}^{N_s\times 3}.
\end{equation}

\paragraph{Decoupled Detail Fields.}
We model high-frequency details with two displacement fields on the dense mesh:
\[
\Delta_g^{s} \in \mathbb{R}^{N_s\times 3},
\qquad
\Delta_f^{s}(i) \in \mathbb{R}^{N_s\times 3}.
\]
Here, $\Delta_g^{s}$ is a global per-vertex, identity-dependent offset that captures expression-invariant geometric details, while $\Delta_f^{s}(i)$ captures expression-dependent facial details for frame $i$. To make the decoupled detail fields optimizable even from a single reference image, we constrain $\Delta_g^{s}$ to mainly deform the non-facial regions (e.g., hair and clothing regions), and $\Delta_f^{s}(i)$ to affect only the facial region. In this way, the detailed canonical mesh for frame $i$ is:
\begin{equation}
\label{eq:displaced_meshes}
\widetilde{V}^{s}(i) \;=\; V^{s} \;+\; \Delta_g^{s} \;+\; \Delta_f^{s}(i).
\end{equation}

\paragraph{Posing and Projection.}
Let $\mathcal{T}_i$ denote the SMPL-X Linear Blend Skinning (LBS) operator at frame $i$, which applies a rigid transformation to each vertex based on its skinning weights. We obtain dense skinning weights for the high-resolution mesh by barycentrically interpolating the base SMPL-X weights. LBS is then applied to the detailed canonical vertices:
\begin{equation}
\label{eq:lbs_mesh}
\widetilde{V}^{s}_{p}(i) \;=\; \mathcal{T}_i\!\left(\,\widetilde{V}^{s}(i)\,\right) \;\in\; \mathbb{R}^{N_s\times 3}.
\end{equation}

Finally, given camera parameters $\mathbf{c}$, we project the posed vertices to the image plane using the projection operator $\Pi$.

\subsubsection{Optimization}
To supervise the 3D representation in image space, we render geometric buffers from the posed dense mesh. A differentiable renderer $\mathcal{R}$~\cite{laine2020modular} takes $\widetilde{V}^{s}_{p}(i)$ and camera $\mathbf{c}$ as input, and outputs per-frame renderings:
\[
(\hat{N}_i,\;\hat{D}_i) \;=\; \mathcal{R}\!\left(\widetilde{V}^{s}_{p}(i),\,\mathbf{c}\right),
\]
where $\hat{N}_i$ is the rendered normal map and $\hat{D}_i$ is the rendered depth map.

\noindent\textbf{Supervision Signals.}
We impose sparse landmarks and pixel-wise dense labels as supervision signals. Let $L_{\textrm{3D}}(i)$ be the selected landmarks from the posed mesh, the landmark loss is defined as:
\begin{equation}
\label{eq:loss_ldmk}
\mathcal{L}_{\textrm{ldmk}}
\;=\;
\big\|\Pi(L_{\textrm{3D}}(i),\,\mathbf{c})-L_{\textrm{2D}}(i)\big\|_2^2,
\end{equation}
where $L_{\textrm{2D}}(i)$ is detected with a keypoint detector~\cite{khirodkar2024sapiens}. For dense supervision, we compare the rendered predictions $(\hat{N}_i,\hat{D}_i)$ against estimated targets $(N_i,D_i)$ from~\cite{saleh2025david}:
\begin{equation}
\label{eq:loss_dense}
\mathcal{L}_{\textrm{normal}} \;=\; \|\hat{N}_i - N_i\|_1,
\qquad
\mathcal{L}_{\textrm{depth}} \;=\; \|\hat{D}_i - D_i\|_1.
\end{equation}

\noindent\textbf{Geometric Regularization.}
We apply an $\ell_2$ prior on the per-frame expression coefficients:
\begin{equation}
\label{eq:loss_exp}
\mathcal{L}_{\textrm{exp}} \;=\; \|\psi_i\|_2^2.
\end{equation}

To ensure physical plausibility and suppress high-frequency artifacts, we also apply a displacement-magnitude penalty $\mathcal{L}_{\textrm{dis}}$ and a Laplacian smoothness term $\mathcal{L}_{\textrm{lap}}$ on both detail fields. The total loss $\mathcal{L}$ is a weighted sum of the above terms.

\subsubsection{Disentanglement Principle}
\label{disentanglement}
A key challenge in our formulation is that the decomposition in \cref{eq:displaced_meshes} is ill-posed since static identity details could be ambiguously baked into either $\Delta_g^s$ or all $\Delta_f^s(i)$ fields. Therefore, we use spatial constraints (facial vs. non-facial) as the initial prior, and on this basis, we introduce an explicit temporal regularizer to force disentanglement and prevent identity information from being leaked into the frame-by-frame dynamics.

Our guiding principle is that for a given video sequence, the static field $\Delta_g^s$ should capture the \textit{average} personalized geometry, while the dynamic fields $\Delta_f^s(i)$ should represent the \textit{minimal, zero-mean deviations} from that average. To enforce this, the $\Delta_g^s$ is initialized to be non-zero in expected static regions as a prior, and we apply a minimal magnitude penalty to each per-frame offset $\Delta_f^s(i)$. This encourages the optimizer to explain all static and shared geometry using the shared $\Delta_g^s$ field, effectively forcing $\Delta_g^s$ to represent the true geometric average, while $\Delta_f^s(i)$ models only the expressive dynamics.

\subsection{Expression Transfer Module}
\label{sec:expression}
As shown in \cref{sec:personalized}, personalized head details are learned in an identity-specific manner. Directly applying the expression offsets of one identity to another (even after alignment) leads to identity-expression mismatch. For example, a child should not inherit the deep wrinkle patterns of an elderly subject. To address this, we introduce an identity-adaptive expression transfer module that renders the same driving expression in a way that is anatomically consistent for the target identity, avoiding artifacts and ambiguity. 

\begin{figure}[t]
\centering
\includegraphics[width=0.8\linewidth]{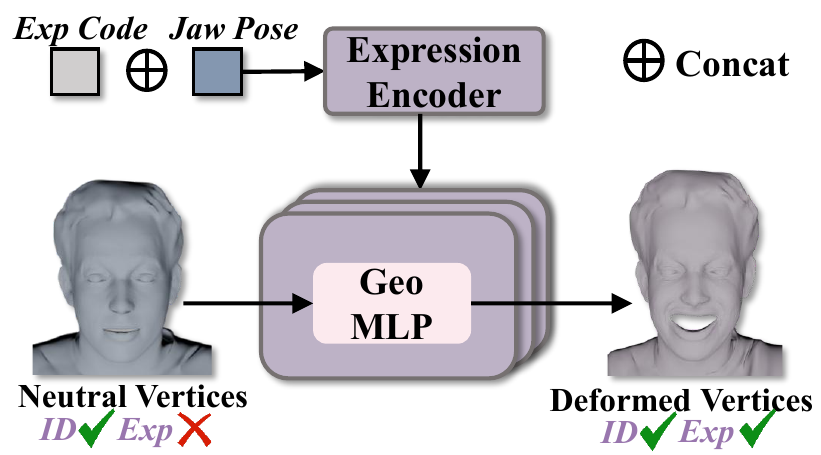}
\caption{Design of our Expression Transfer Module.}
\label{fig:Expression-Transfer}
\end{figure}

As shown in \cref{fig:Expression-Transfer}, the module has two parts: (i) an expression encoder that encodes driving signals, and (ii) a vertex-wise geometry MLP that predicts per-vertex dynamic offsets conditioned on the encoded expression and the target identity geometry.

\paragraph{Driving Signals Encoder.}
Let the driving expression coefficients over $F$ frames be $\boldsymbol{\psi} \in \mathbb{R}^{F \times 100}$ and the jaw poses be $\boldsymbol{\omega} \in \mathbb{R}^{F \times 3}$. We feed them into an encoder $\mathcal{E}$ to obtain per-frame conditioning codes:
\begin{equation}
\label{eq:expr_encoder}
Q \;=\; \mathcal{E}\!\big(\boldsymbol{\psi},\,\boldsymbol{\omega}\big) \;\in\; \mathbb{R}^{F \times D},
\end{equation}
where $Q=\{q_i\}_{i=1}^{F}$ and $q_i \in \mathbb{R}^{D}$ summarizes the driving motion at frame $i$.

\paragraph{Details Prediction Network.}
We use a lightweight MLP to predict expression-dependent per-vertex offsets and add them to the static identity mesh. Consistent with \cref{sec:personalized}, let
\[
V_{\textrm{neutral}} \;=\; V^{s} + \Delta_g^{s} \;\in\; \mathbb{R}^{N_s \times 3},
\]
where $V^{s}$ is the dense canonical mesh and $\Delta_g^{s}$ is the static, identity-dependent offset. The dynamic offsets are predicted by:
\begin{equation}
\label{eq:mlp_offsets_simple}
\Delta_f^{s}(i) \;=\; \mathcal{G}\!\big(V_{\textrm{neutral}},\, q_i\big) \;\in\; \mathbb{R}^{N_s \times 3}.
\end{equation}

By predicting dynamic offsets conditioned on both the driving code $q_i$ and the target identity’s neutral geometry, the module transfers expressions across identities while adapting high-frequency details (e.g., wrinkle placement and intensity) to the target’s anatomy, resolving identity-expression incompatibilities and yielding robust expression transfer.

\begin{figure*}[ht]
\centering
\includegraphics[width=\linewidth]{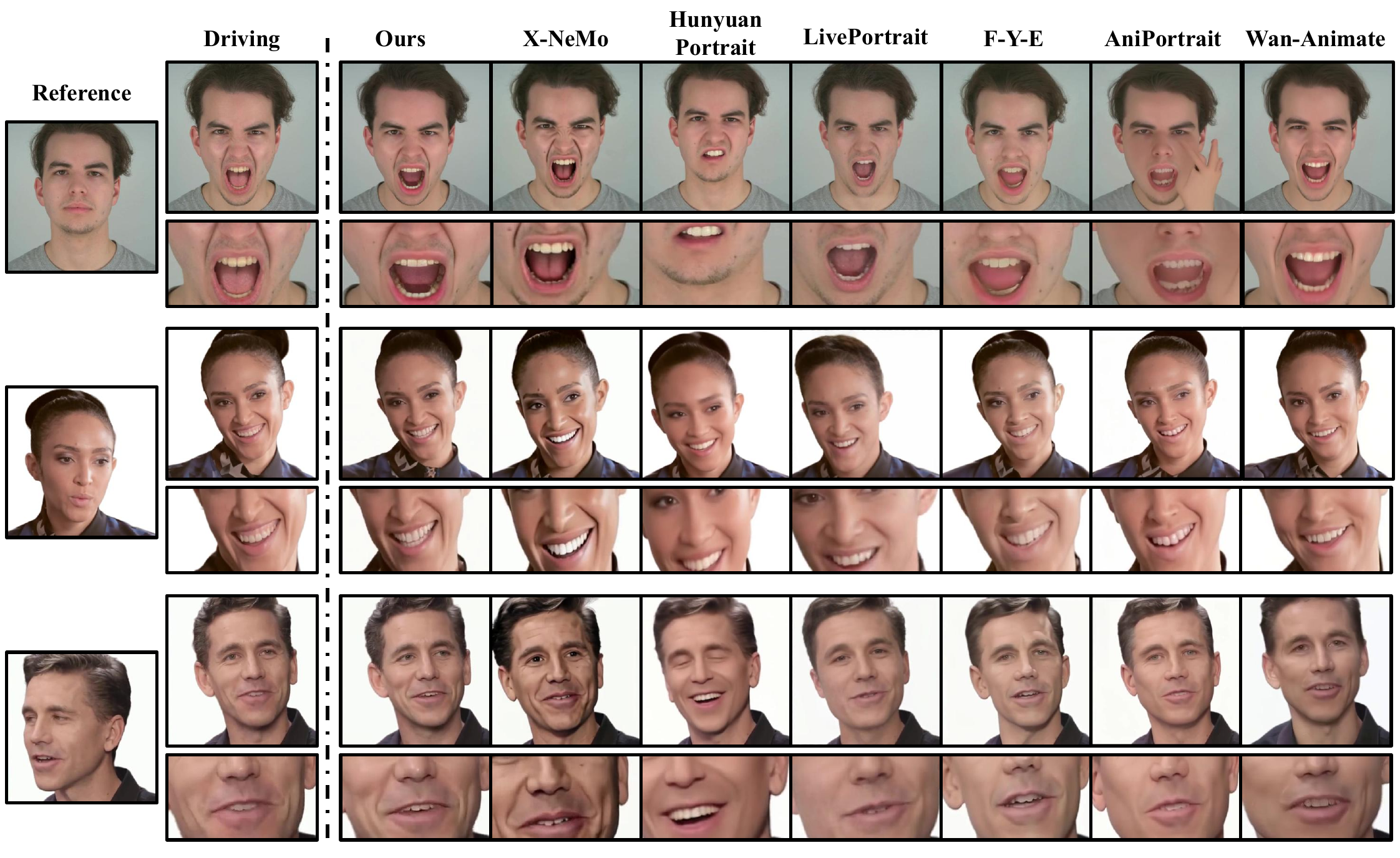}
\caption{Qualitative results in self-reenactment. Compared to other methods, our method can reveal more details about identity and facial expressions.}
\label{fig:qualitative_self}
\end{figure*}

\begin{figure*}[ht]
\centering
\includegraphics[width=\linewidth]{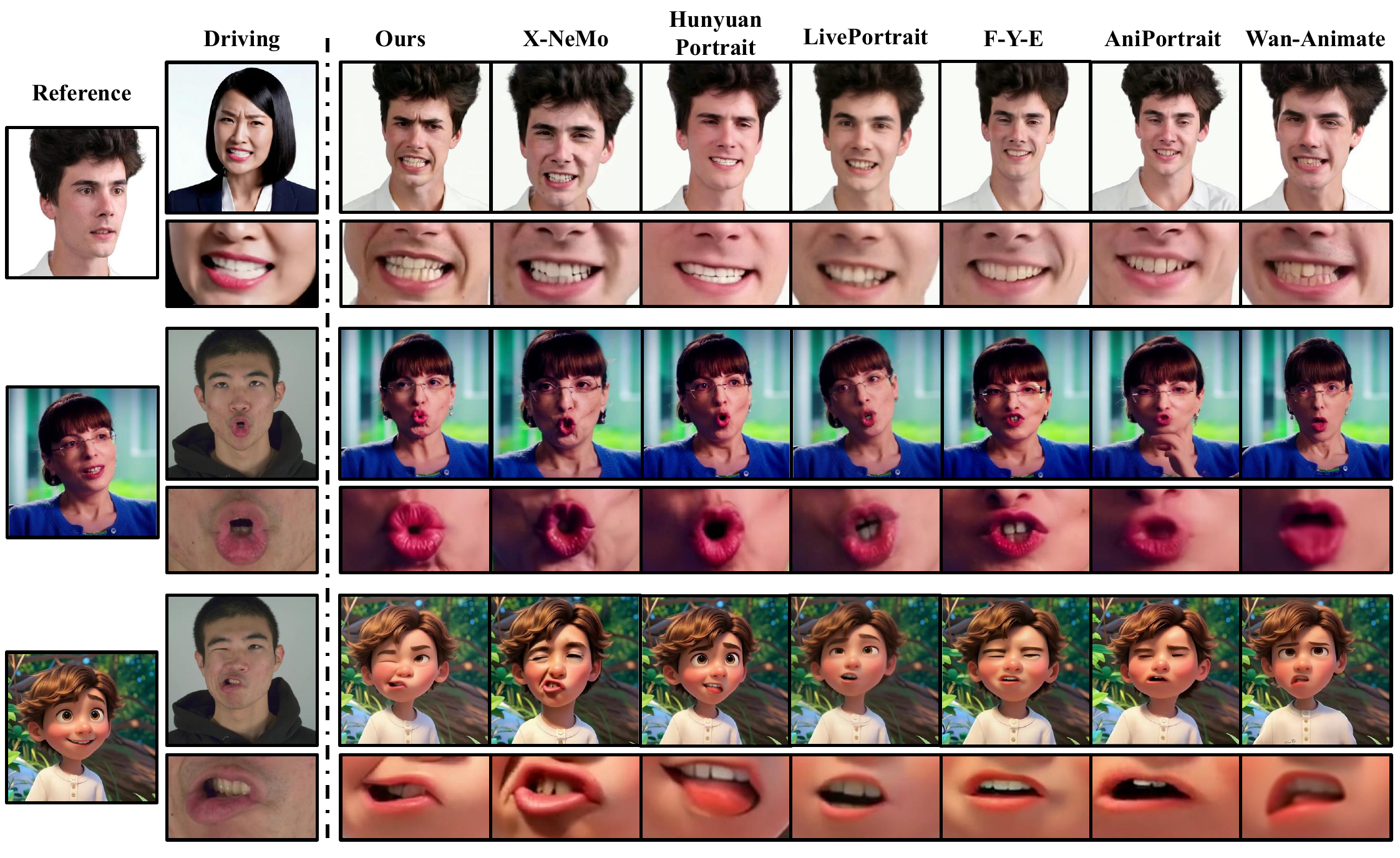}
\caption{Qualitative results in cross-reenactment. Our method can effectively transfer body posture and facial expressions while ensuring identity consistency.}
\label{fig:qualitative_cross}
\end{figure*}

\subsection{Video Diffusion Model}
\label{sec:diffusion}
To demonstrate the utility of our head representation, we fine-tune a pretrained video generation model~\cite{wan2025wan}, implemented as a Diffusion Transformer (DiT)~\cite{peebles2023scalable} in the latent diffusion framework (LDM)~\cite{rombach2022high}.

\paragraph{Control Signals.}
Given the reference image $I_R$, we build the personalized neutral head mesh $V_{\text{neutral}}$ (\cref{sec:personalized}) and render a reference normal map:
\[
N^{R} \;=\; \mathcal{R}\!\left(V_{\text{neutral}},\,\mathbf{c}\right).
\]
For the driving sequence $I_D=\{I_{D_i}\}_{i=1}^{F}$, we use the expression transfer module (\cref{sec:expression}) to obtain $\widetilde{V}^{s}(i)$ for each frame $i$, pose it with LBS to $\widetilde{V}^{s}_{p}(i)$, and render the corresponding driving normal maps:
\[
N^{D}_{1:F} \;=\; \big\{\,\mathcal{R}\!\left(\widetilde{V}^{s}_{p}(i),\,\mathbf{c}\right)\big\}_{i=1}^{F}.
\]

\paragraph{Conditioned Denoising.}
Following~\cite{hu2024animate,wang2025unianimate,zhang2025mimicmotion}, a 3D convolutional pose encoder extracts spatio-temporal features from $N^{D}_{1:F}$, while a 2D convolutional reference encoder extracts appearance cues from $N^{R}$. Let $z_0$ denote the clean video latent from the Causal VAE~\cite{wan2025wan} and $z_t$ its noised version at timestep $t$. We patchify the control features from $N^{D}_{1:F}$ and concatenate them with the patchified tokens of $z_t$ (together with the reference features from $N^{R}$) to form the conditioning input $c$ for the DiT backbone.

We adopt the standard LDM noise-prediction loss. The DiT denoiser $\epsilon_\theta$ predicts the added noise $\epsilon$:
\begin{equation}
\mathcal{L}_{\textrm{ldm}} \;=\; \mathbb{E}_{z_0,\,\epsilon \sim \mathcal{N}(0,1),\,t}\!\left[\,\big\|\epsilon - \epsilon_\theta(z_t,\,t,\,c)\big\|_2^2\,\right],
\end{equation}
where $c$ encodes the control signals described above. In the cross-driving setting, $N^{D}_{1:F}$ is rendered from the expression-transferred meshes derived by combining the driving SMPL-X expression parameters with the reference identity’s personalized neutral mesh, ensuring identity-consistent yet expressive guidance.

%% file: sec/4_experiments.tex
\section{Experiments}
\label{sec:Experiments}
\subsection{Implementation Details}
\label{dataset}

\begin{table*}[t]
\centering
\caption{Quantitative comparison. Our method achieves superior numerical results to all the baselines in self- and cross-reenactment tasks, evaluated on an image resolution of $512\times 512$.}
\label{tab:quant_rec}


\small 
\setlength{\tabcolsep}{3pt} 

{
\begin{tabular}{lccccccc|ccc}
\toprule

\multirow{2}{*}{Method} & \multicolumn{7}{c|}{\textbf{Self Reenactment}} & \multicolumn{3}{c}{\textbf{Cross Reenactment}} \\

\cmidrule(lr){2-8} \cmidrule(lr){9-11}

& \textbf{PSNR} $\uparrow$ & \textbf{SSIM} $\uparrow$ & \textbf{LPIPS} $\downarrow$ & \textbf{L1} $\downarrow$ & \textbf{AED} $\downarrow$ & \textbf{APD} $\downarrow$ & \textbf{CSIM} $\uparrow$ & \textbf{AED} $\downarrow$ & \textbf{APD} $\downarrow$ & \textbf{CSIM} $\uparrow$ \\
\midrule

LivePortrait~\cite{guo2024liveportrait} &23.2897 &0.82985 &0.37339 &0.04619 & \textbf{0.12924} & 0.02144 & 0.82972 &0.28641 &0.23048 & \underline{0.72869} \\
AniPortrait~\cite{wei2024aniportrait} &22.2527 &0.80838 &0.30018 &0.04078 & 0.17581 & 0.01631 & 0.76038 &0.25256 &0.02161 & 0.62998 \\
Follow-Your-Emoji~\cite{ma2024follow} &25.6872 &0.84146 &0.23645 &0.02905 & 0.14717 & 0.01485 & 0.80284 &0.22069 &0.02338 &0.68373 \\
Hunyuan Portrait~\cite{xu2025hunyuanportrait} &22.9453 &0.79333 &0.27573 &0.03962 & 0.15593 & 0.02329 & 0.81150 &0.22753 &0.08784 & 0.67451 \\
X-NeMo~\cite{zhao2025xnemo} &21.5618 &0.78051 &0.32411 &0.04760 & 0.13671 & 0.01789 & \underline{0.83018} &\textbf{0.17092} &0.02101 &0.72164 \\
Wan-Animate~\cite{cheng2025wan} &\underline{26.5109} &\underline{0.84360} &\underline{0.21397} &\underline{0.02183} & 0.14415 & \underline{0.00955} & 0.82566 & 0.22156 & \underline{0.01745} & 0.72637 \\
\midrule
Ours &\textbf{26.5507} &\textbf{0.85908} &\textbf{0.18394} &\textbf{0.02160} & \underline{0.13185} & \textbf{0.00948} & \textbf{0.83506} &\underline{0.21092} &\textbf{0.01277} & \textbf{0.72917} \\
\bottomrule
\end{tabular}
}
\end{table*}

We collect and process 4{,}000 videos from VFHQ~\cite{xie2022vfhq}, CelebV-HQ~\cite{zhu2022celebv}, and HDTF~\cite{zhang2021flow} datasets, totaling approximately 10 hours of footage. All frames are cropped and resized to $512\times512$. We then apply SMPL-X reconstruction and the joint geometric optimization described in \cref{sec:personalized} to obtain personalized head models. The expression parameters and personalized meshes extracted from all videos are used to train the identity-adaptive expression transfer module. Subsequently, the full set of personalized, detailed head representations together with the original videos serves as training data for the diffusion model, during which the expression transfer module is frozen. We train for 30 epochs on $4\times$ NVIDIA A800 GPUs with a batch size of 1 per GPU and a learning rate of $10^{-4}$.

Given the stringent requirements of film-grade digital humans regarding identity consistency, expressive fidelity, and high visual clarity, we conduct a fair evaluation on two datasets that are not used for training: RAVDESS~\cite{livingstone2018ryerson} and NeRSemble~\cite{kirschstein2023nersemble}. These datasets feature rich facial expressions and high video quality. The RAVDESS test set includes 20 portrait videos, and the NeRSemble test set comprises 80 portrait videos. At inference, we sample frames with a temporal stride to assess generalization to head poses and expressions that differ from the reference image. All metrics are computed at $512\times512$ resolution.

\subsection{Evaluations and Comparisons}
\label{Comparison}

\begin{figure}[t]
\centering
\includegraphics[width=\linewidth]{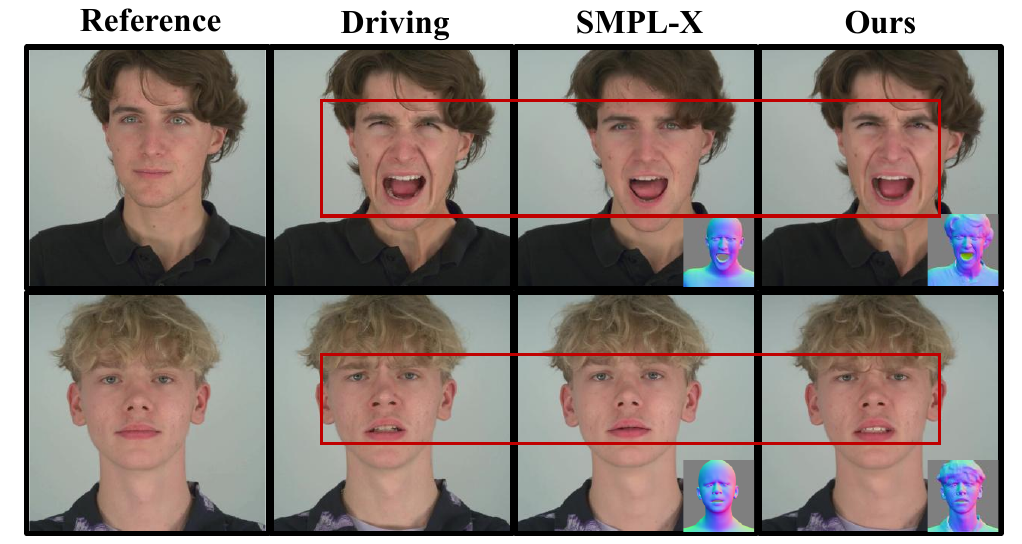}
\caption{Ablation study between SMPL-X and our head representation.}
\label{fig:ablation_1}
\end{figure}

To demonstrate the effectiveness of our method, we compare against prior portrait generation methods, including \textit{LivePortrait}~\cite{guo2024liveportrait}, an implicit keypoint-based approach; \textit{AniPortrait}~\cite{wei2024aniportrait} and \textit{Follow-Your-Emoji (F-Y-E)}~\cite{ma2024follow}, which use FLAME- and keypoint-driven explicit control; \textit{X-NeMo}~\cite{zhao2025xnemo}, which relies on implicit latent feature control from a driving image; as well as \textit{Wan-Animate}~\cite{cheng2025wan} and \textit{Hunyuan Portrait}~\cite{xu2025hunyuanportrait}, which utilize explicit pose representations and implicit expression features for control.

Following \textit{Learn2Control}~\cite{gao2025constructing}, we conduct both qualitative and quantitative evaluations to assess video quality and motion accuracy. For self-reenactment, we report PSNR, SSIM, LPIPS~\cite{zhang2018unreasonable}, and $\ell_1$ distance. Identity preservation is measured by cosine similarity (CSIM) between face recognition embeddings~\cite{deng2019arcface}. Motion accuracy is quantified by the Average Expression Distance (AED) and Average Pose Distance (APD). For cross-reenactment, we report AED, APD, and CSIM.


\noindent \textbf{Quantitative Comparison.}
\label{result analysis}
As demonstrated in \cref{tab:quant_rec}, our method achieves state-of-the-art performance, quantitatively outperforming all baselines in both self- and cross-reenactment tasks. In self-reenactment, our approach produces results with significantly higher reconstruction fidelity while more accurately preserving the target's pose and identity. For the more challenging cross-reenactment task, our method demonstrates a superior ability to faithfully transfer motion and expression while robustly preserving the source subject's identity.

\noindent \textbf{Qualitative Analysis.}
As illustrated in the qualitative comparisons (\cref{fig:qualitative_self,fig:qualitative_cross}), existing methods exhibit distinct limitations that underscore the necessity of our personalized head representation. Explicit-control baselines, such as \textit{AniPortrait} and \textit{F-Y-E}, rely on sparse, low-rank parametric signals. These signals are insufficient for capturing fine-grained details, resulting in animations with limited expressiveness and weaker identity preservation. In contrast, implicit-control approaches, including \textit{LivePortrait}, \textit{Hunyuan Portrait}, and \textit{X-NeMo}, utilize latent motion features to achieve greater expressiveness. However, this paradigm introduces severe trade-offs: the control over head pose is often imprecise, and the motion features frequently entangle expression dynamics with identity attributes. This critical entanglement leads to identity leakage in both self-reenactment (via identity drift) and cross-reenactment (via identity bleed-through from the driver). Notably, while \textit{Wan-Animate} demonstrates that simply scaling up data and model parameters can enhance generalization capabilities, it fails to significantly improve fine-grained controllability or the decoupling of identity and expression. As shown in \cref{fig:qualitative_cross}, \textit{Wan-Animate} still falls short in precise expression control and identity decoupling. Benefiting from our personalized and disentangled 3D head representation, our \textit{ExpPortrait} effectively overcomes these issues. It successfully decouples identity from expression, enabling the simultaneous generation of high-fidelity, high-expressiveness, and highly-controllable portrait animations that faithfully preserve identity while accurately reproducing nuanced expressions.

\subsection{Ablation Studies}
\label{ablation}
We assess the efficacy of our personalized head representation via an ablation against a SMPL-X baseline. The baseline uses the standard SMPL-X mesh as the sole 3D head representation, whereas our full model employs the proposed personalized, high-detail head proxy. For a fair comparison, both variants share the same diffusion architecture, training data, and hyperparameters; the only difference is the underlying 3D head representation. As shown in \cref{fig:ablation_1}, the SMPL-X baseline struggles to capture fine facial nuances and exhibits limited facial dynamics, often producing rigid or muted expressions. In contrast, our representation delivers markedly better expressiveness and identity fidelity, recovering a wider and more accurate range of personalized identity cues and subtle expression details. 

\begin{figure}[t]
\centering
\includegraphics[width=\linewidth]{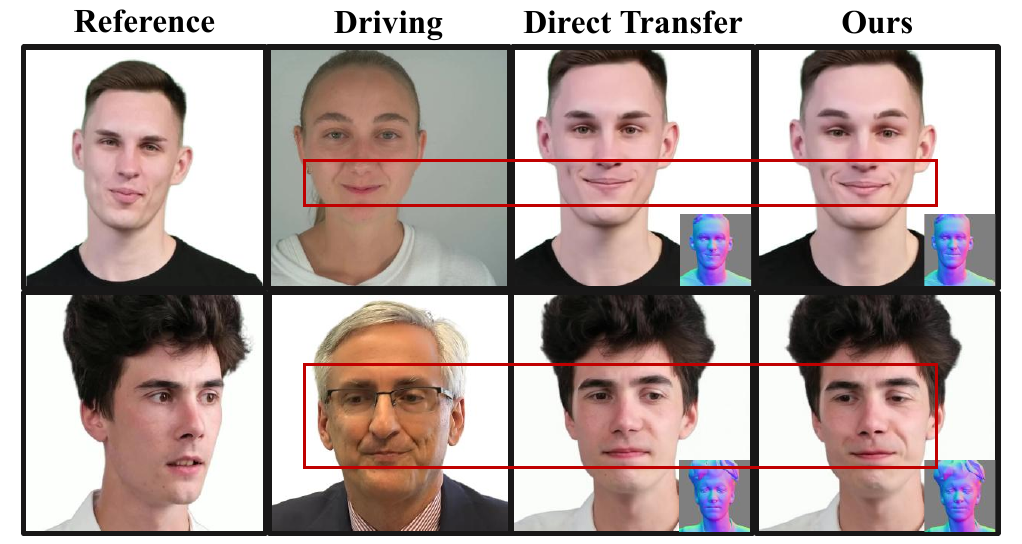}
\caption{Ablation study between directly performing deformation with offsets and using our expression transfer module.}
\label{fig:ablation_2}
\end{figure}

We also conduct an ablation study on our expression transfer module, benchmarking it against a strategy that directly applies expression offsets to perform deformation on the reference subject's head representation. As illustrated in \cref{fig:ablation_2}, this direct transfer approach exhibits significant drawbacks: it yields muted facial expressiveness. In contrast, our proposed expression transfer module is capable of generating more vivid and realistic expressions.

%% file: sec/5_conclusion.tex
\section{Conclusion}
In this work, we addressed the problem of expressive, controllable, and identity-preserving portrait video generation. We introduced a high-fidelity personalized head representation that disentangles identity-specific static geometry from frame-wise expression details, and an identity-adaptive expression transfer module. Conditioned on this 3D proxy, a DiT–based generator achieved state-of-the-art performance, leading to clear improvements in identity preservation and expression accuracy.

Our method still has limitations. The personalized representation does not explicitly model the inner mouth, hindering the precise generation of the tongue, nor does it capture fine-grained eyeball motion. Nevertheless, we believe this representation provides a promising foundation for natural extensions to full-body animation and other tasks, such as speech-driven animation and broader interactive avatar applications, which we leave for future investigation.